\documentclass[conference]{IEEEtran}
\IEEEoverridecommandlockouts 

\usepackage{amsmath}
\usepackage{amssymb}
\usepackage[nolist]{acronym}
\usepackage{bm}
\usepackage{color}
\usepackage{cite}
\usepackage[]{graphicx}

\graphicspath{{./figures}}

\usepackage{titlesec}
\titlespacing*{\section}
{0pt}{6px}{6px}
\titlespacing*{\subsection}
{0pt}{6px}{2px}

\newacro{dof}[DoF]{Degrees of Freedom}
\newacro{dl}[DL]{Deep Learning}
\newacro{ds}[DS]{Dynamical Systems}
\newacro{gmr}[GMR]{Gaussian Mixture Regression}
\newacro{ik}[IK]{Inverse Kinematics}
\newacro{il}[IL]{Imitation Learning}
\newacro{ilvs}[ILVS]{Imitation Learning Visual Servoing}
\newacro{nn}[NN]{artificial Neural Network}
\newacro{vs}[VS]{Visual Servoing}
\newacro{rds}[RDS]{Reshaped \acl{ds}}
\newacro{clfdm}[CLF-DM]{Control Lyapunov Function-based Dynamic Movements}
\newacro{clf}[CLF]{Control Lyapunov Function}
\newacro{fdm}[FDM]{Fast Diffeomorphic Matching}
\newacro{dmp}[DMP]{Dynamic Movement Primitive}
\newacro{seds}[SEDS]{Stable Estimator of Dynamical Systems}

\newcommand{\FriW}[1]{\emph{FriWalk}}

\begin{document}

\title{\LARGE \bf
Human Motion Prediction for Human-Robot Collaboration
}

\author{Placido Falqueto$^{1,*}$, Elena Basei$^{1,*}$, Edoardo Lamon$^1$, Giovanni Perantoni$^1$, \\ Matteo Saveriano$^2$, Daniele Fontanelli$^2$, Luigi Palopoli$^{1}$
\thanks{$^{*}$Authors contributed equally to this work}%
\thanks{$^{1}$Dept. of Information Engineering and Computer Science (DISI), University of Trento, Trento, Italy {\tt\small \{name.surname\}@unitn.it}}%
\thanks{$^{2}$Department of Industrial Engineering (DII), University of Trento, Trento, Italy {\tt\small \{name.surname\}@unitn.it}}%
}

\maketitle
\thispagestyle{empty}
\pagestyle{empty}

\begin{abstract}
This paper compares three human motion prediction methods — GMM-based clustering, TransFusion, and Graph-Mixer — evaluating their effectiveness in human-robot collaboration. These predictions are valuable for improving robot motion planning by anticipating human movements and enhancing safety and efficiency in dynamic environments.
\end{abstract}



\section{Introduction}\label{sec:intro}

Predicting human motion in real-time is key to safe human-robot collaboration. Traditional approaches, like Gaussian Mixture Models (GMMs)\cite{MainpriceGMM}, work in controlled scenarios but struggle with complex human behaviour. Deep learning (DL) methods, including RNNs, GCNs, and Transformers, address these limitations by improving prediction accuracy\cite{YangAdaptive,TianTransF}, though the real-time application is challenging due to computational demands.

Traditional motion planning methods ensure safety but lack adaptability for dynamic interaction. Proactive approaches anticipate human actions, with methods like temporal PRM enhancing collaboration~\cite{HuppiTPRM}. Our work compares clustering and deep learning for HMP, aiming to integrate predictions into a human-aware planning framework for improved safety and responsiveness.
\section{Methodology}\label{sec:method}

We present a comparison of real-time approaches to human motion prediction.
One approach is to use clustering techniques to group similar gestures. The human movements are segmented based on variations in the hands' velocity vectors, which often indicate transitions between gestures in a repetitive industrial setting. Once segmented, gestures are clustered using a Gaussian Mixture Model (GMM) with Dynamic Time Warping (DTW) and Principal Component Analysis (PCA) for time-series comparison.

We need to create clusters that are ($T \times J \times 3$)-dimensional, where $T$ is the number of time samples and $J$ is the number of human joints considered $\times 3$ to consider the 3D cartesian positions. To reduce the dimensionality of the problem, hence increasing real-time performances, we used PCA to preprocess all data, bringing the problem to $T \times N$ dimensions, where $N$ is the number of components we decided to keep after PCA.

\begin{figure}[h]
    \centering
    \includegraphics[width=0.35\linewidth]{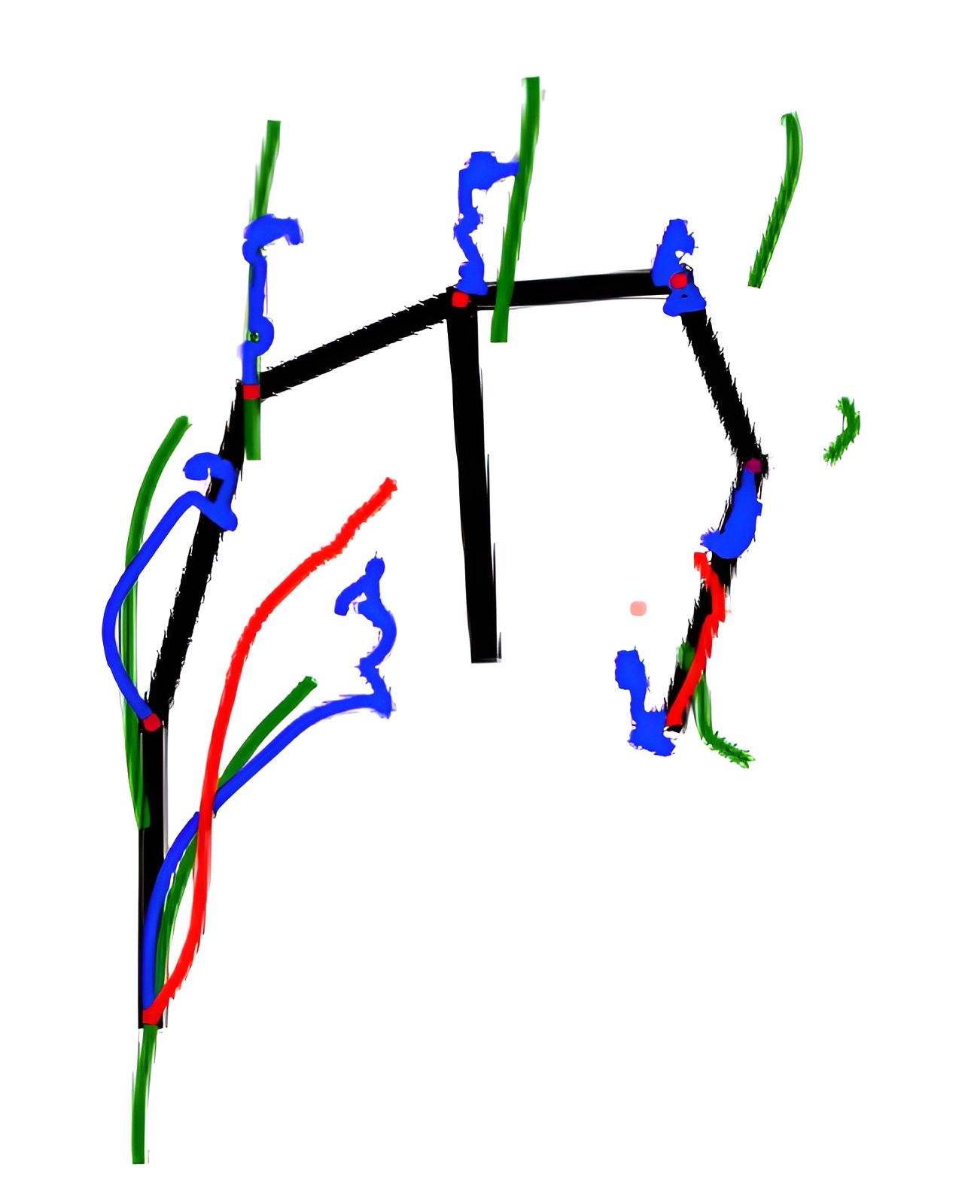}
    \caption{Comparison of predictions using the clustering method and deep learning method for the prediction. The red trails are the ground-truths of the joint trajectories of the human. The green trails are the predictions using the clustering method. The blue trails are the predictions using the deep learning method.}
    \label{fig:comp_pred}
\end{figure}

DTW allows us to generalize a path to multiple trajectories with different profiles of speed of execution.
This lowers the burden on the clustering model since the same cluster can represent many trajectories that follow the same geometric path.

We tried both random sampling and K-Means clustering to initialize the GMM clusters. The K-Means initialization reached a slightly lower prediction error, so we decided to implement it over the random initialization.
After the initialization, an Expectation–maximization algorithm (EM) is used to converge to the final multivariate Gaussians' parameters.
The centroids of the closest clusters are then used for the motion prediction inference in real-time.
The clusters are retrieved with their responsibilities and probability values of being the Gaussian that generates the current gesture.

Alongside clustering, we investigate DL approaches. Specifically, we employ a graph-based model (Graph-Mixer), which consists of an initial pose embedding module that processes the input pose sequences into higher-dimensional features using an adaptive spatial graph convolution. This is followed by the spatial-temporal graph-mixer, which incorporates both adaptive spatial and temporal graph convolutional networks to model human skeleton dynamics and temporal relationships across different frames. Finally, the prediction head generates future motion predictions from these spatial-temporal features, allowing for more accurate long-term human motion forecasting. Additionally, we explore TransFusion, a Transformer-based architecture combined with a diffusion process to predict 3D human motion. The model uses long skip connections between shallow and deep layers to enhance information flow and employs squeeze-and-excitation (SE) blocks for token recalibration. Inputs, including historical motion data and diffusion steps, are treated as tokens. The model operates in the frequency domain using a discrete cosine transform (DCT) to process motion data, reducing noise and dimensionality. 

We integrate real-time human predictions into the robot’s planning process. The robot continuously adapts its motion based on predicted human movements, ensuring safe and smooth collaboration. This dynamic adjustment allows the robot to anticipate human actions and modify its path accordingly.


\section{Results and Discussion}\label{sec:experiments}
This section presents the results and compares the three prediction methods: the Clustering-Based prediction method, the Adaptive Spatial-Temporal Graph-Mixer network, and the TransFusion network. We trained and tested our system using the HA4M dataset~\cite{CicirelliHA4M}. 

\subsection{Training of the Models}
\label{subsec:training}
For the GMM approach, we decided to keep 10 components after the PCA, and create $100$ clusters. These values were chosen after an ablation study.
The length of the segmented gestures is variable and handled by the DTW, but corresponds roughly to 2 seconds of human joint trajectories.
We decided to use 6 frames of historical motion to predict 60 frames (2 seconds) of future motion for the deep learning (DL) approaches. The Graph-Mixer network was trained for 20 epochs, while the TransFusion network was trained for 500 epochs.

\subsection{Evaluation of Human Motion Prediction Models}
\label{subsec:prediction_results}

A qualitative evaluation and comparison of the GMM and TransFusion predictions is shown in Figure~\ref{fig:comp_pred}.
We evaluated our methods using the Mean Per Joint Position Error (MPJPE), a standard metric in 3D human pose prediction. This metric calculates the average Euclidean distance between the predicted and actual joint positions in 3D space. The MPJPE is calculated by computing the error for each joint, averaging across all joints, and then averaging across all frames in a sequence. Formally:

\begin{equation}
    MPJPE = \frac{1}{N}\sum_{i=1}^{N}\lVert \hat{J}_{i} - J_{i} \rVert
\end{equation}

where $N$ is the number of joints, $\hat{J}_{i}$ is the predicted 3D position of the i-th joint, $J_{i}$ is the ground truth 3D position of the i-th joint, and $\lVert \cdot \rVert$ denotes the Euclidean distance between the predicted and ground truth joint positions. 
In Table \ref{table:comparison} we report the MPJPE, variance, the 5th percentile, and the 95th percentile results for the three models.




\begin{table}[h!]
\centering
\caption{Method comparison}
\begin{tabular}{ c | c | c | c | c}
    Method & MPJPE & Variance & 5th Percentile & 95th Percentile\\
    \hline
    Clustering & $71.60$ & $2.14$ & $7.13$ & $182.25$ \\
    TransFusion	& $100.81$ & $5853.17$ & $20.57$ & $251.47$ \\
    Graph-Mixer & $111.27$ & $10381.51$ & $14.44$ & $343.57$ \\
\end{tabular}
Note: All values are in millimeters [mm]
\label{table:comparison}
\end{table}


\subsection{Prediction-aware motion planning}\label{subsec:discussion}
Our preliminary simulations indicate that integrating human motion prediction into robot planning using the t-PRM~\cite{HuppiTPRM} leads to collision-free trajectories for the manipulator. Such trajectories improve safety and efficiency in performing an industrial task.
\section{Conclusions}\label{sec:conclusion}

In this work, we validated and compared three different
human motion prediction approaches: the Clustering-Based GMM prediction method, the Adaptive Spatial-Temporal Graph-Mixer network, and the TransFusion network. Our preliminary simulations validated the quality of the predictions of
both traditional clustering and modern deep learning techniques, and
demonstrated a good capability of avoiding human
motions during robot operation.

For future work, we plan to test the framework experimentally, using the ZED2 camera and a Universal Robots UR5.
We plan to extend our approach by incorporating
probabilistic scenario-based planning
techniques~\cite{OleinikovScenario}, to enhance our system's ability
to handle multiple predictions with associated probabilities, since our Clustering-Based prediction method outputs multiple predictions with their corresponding probability. This
extension will enable more robust and adaptive motion planning by
generating and optimizing trajectories based on probabilistic human
motion scenarios.

Our results contribute to advancing human-robot interaction by
improving the robot's proactive collision avoidance capabilities.

\section{Acknowledgments}
Co-funded by the European Union. Views and opinions expressed are
however those of the author(s) only and do not necessarily reflect
those of the European Union or the European Commission. Neither the
European Union nor the granting authority can be held responsible for
them (EU - HE Magician – Grant Agreement 101120731).


\bibliographystyle{IEEEtran}
\bibliography{bibliography}

\end{document}